\documentclass[a4paper, times, 10pt,twocolumn]{article}
\usepackage[top=4.9cm,bottom=3.7cm,left=1.5cm,right=1.5cm]{geometry}
\usepackage{ICMLC}
\usepackage{times}
\usepackage{graphicx}
\usepackage{indentfirst}
\usepackage{latexsym}
\usepackage{algorithm}
\usepackage{algorithmic}
\usepackage{multirow}
\usepackage{graphicx}
\usepackage{float}
\usepackage[section]{placeins}
\usepackage[linewidth=1pt]{mdframed}
\usepackage{graphicx,subfigure}
\usepackage[export]{adjustbox}
\usepackage{caption} 
\usepackage{xcolor}
\usepackage{ulem} 
\usepackage{appendix}
\usepackage{hyperref}
\usepackage{graphicx}
\usepackage{subfigure}
\usepackage{multirow}
\usepackage{tabularx}
\usepackage{booktabs}
\usepackage{adjustbox}
\usepackage{xcolor}
\usepackage{ulem} 
\usepackage{appendix}
\usepackage{hyperref}
\usepackage{graphicx}
\usepackage{subfigure}
\usepackage{amssymb}
\usepackage{multirow}
\usepackage{amsmath}
\usepackage{amssymb}
\usepackage{tabularx}
\usepackage{booktabs}
\usepackage{adjustbox}
\usepackage{caption} 
\usepackage{textcomp}
\usepackage{stfloats}
\usepackage{url}
\usepackage{verbatim}
\usepackage{graphicx}
\hypersetup{
    colorlinks=true,
    linkcolor=blue,
    citecolor=blue,
    urlcolor=blue
}
\definecolor{Sully}{RGB}{85,200,150}
\definecolor{Labrador}{RGB}{255,159,164}
\definecolor{BushJunior}{HTML}{F56EBB}
\definecolor{BushSenior}{HTML}{F0A441}
\definecolor{BarackObama}{HTML}{736E73}
\definecolor{jg}{HTML}{9FC2F5}
\definecolor{mylegal}{RGB}{241, 172, 106}
\usepackage[margin=8pt,font=footnotesize,labelfont=bf,labelsep=period
]{caption}

\pdfpagewidth=\paperwidth
\pdfpageheight=\paperheight
\begin{document}

\title{MUSE: Multi-Knowledge Passing on the Edges, Boosting Knowledge Graph Completion}  

\author{\bf{\normalsize{Pengjie Liu${^1}$}}\\ 
\\
\normalsize{$^1$School of Computer Science and Engineering, Southern University of Science and Technology, Shenzhen, China}\\
\normalsize{E-MAIL: 12031109@mail.sustech.edu.cn}\\
\\}


\maketitle \thispagestyle{empty}
\begin{abstract}
Knowledge Graph Completion (KGC) aims to predict the missing information in the (head entity)-[relation]-(tail entity) triplet. Deep Neural Networks have achieved significant progress in the relation prediction task. However, most existing KGC methods focus on single features (e.g., entity IDs) and sub-graph aggregation, which cannot fully explore all the features in the Knowledge Graph (KG), and neglect the external semantic knowledge injection. To address these problems, we propose MUSE, a knowledge-aware reasoning model to learn a tailored embedding space in three dimensions for missing relation prediction through a multi-knowledge representation learning mechanism. Our MUSE consists of three parallel components: 1) \textit{Prior Knowledge Learning} for enhancing the triplets' semantic representation by fine-tuning BERT; 2) \textit{Context Message Passing} for enhancing the context messages of KG; 3) \textit{Relational Path Aggregation} for enhancing the path representation from the head entity to the tail entity. Our experimental results show that MUSE significantly outperforms other baselines on four public datasets, such as over \textbf{5.50\%} improvement in H@1 and \textbf{4.20\%} improvement in MRR on the NELL995 dataset. The code and all datasets will be released via https://github.com/NxxTGT/MUSE.
\end{abstract}
\section{Introduction}
\label{sec:intro}
\begin{figure}[t!]{
\centering
 \subfigure[\label{fig:1a}
\textbf{Limited Information Set (LIS)}:  Entity Degree $<$ 3. When we predict the relation between the \textbf{\color{Sully}Sully} and \textbf{\color{Labrador}Labrador}, the injected prior knowledge can guide MUSE to identify \textbf{\color{Sully}Sully} is a dog, which should be~\textit{\{Breed}\} of \textbf{\color{Labrador}Labrador} not~\textit{\{Food}\}. 
 ]
 {
            \includegraphics[width=0.4\textwidth]{
            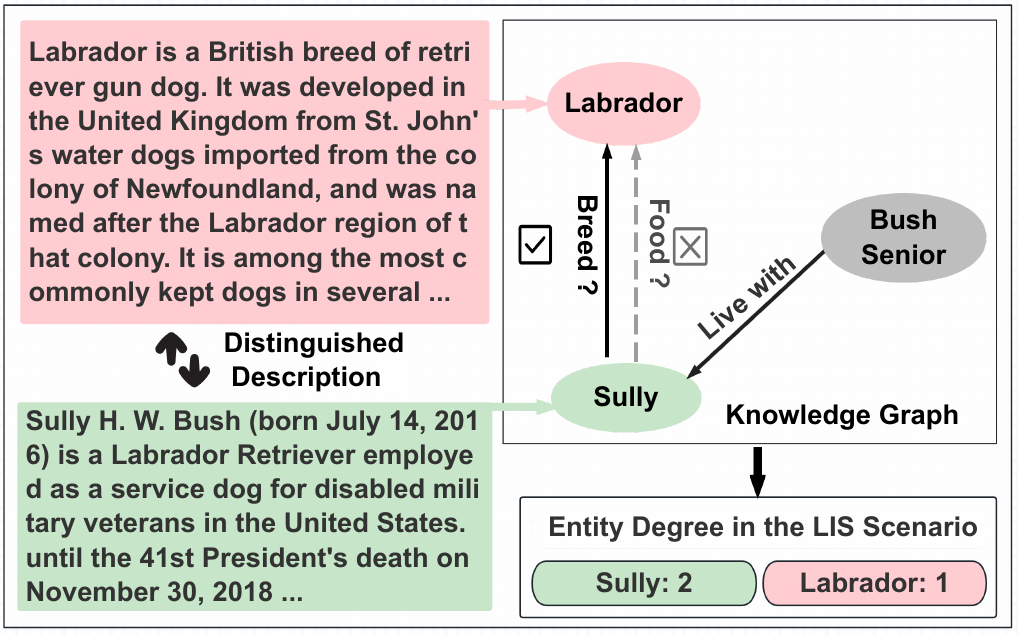
            }
            }
  \subfigure[\label{fig:1b}
\textbf{Rich Information Set (RIS)}: Entity Degree~$\ge$ 3. When predicting \textbf{\color{BushSenior}Bush Senior} as the \textit{\{Father}\} of \textbf{\color{BushJunior}Bush Junior} or \textbf{\color{BarackObama}Barack Obama}, \textbf{\color{BushJunior}Bush Junior} and \textbf{\color{BarackObama}Barack Obama} share similar relations: \textit{\{Mother\}} and \textit{\{President\}}. Their descriptions show many similarities in terms of their presidential terms and political careers. Then the knowledge in contextual and relational paths further enhance their representation.
  ]{
           \includegraphics[width=0.4\textwidth] {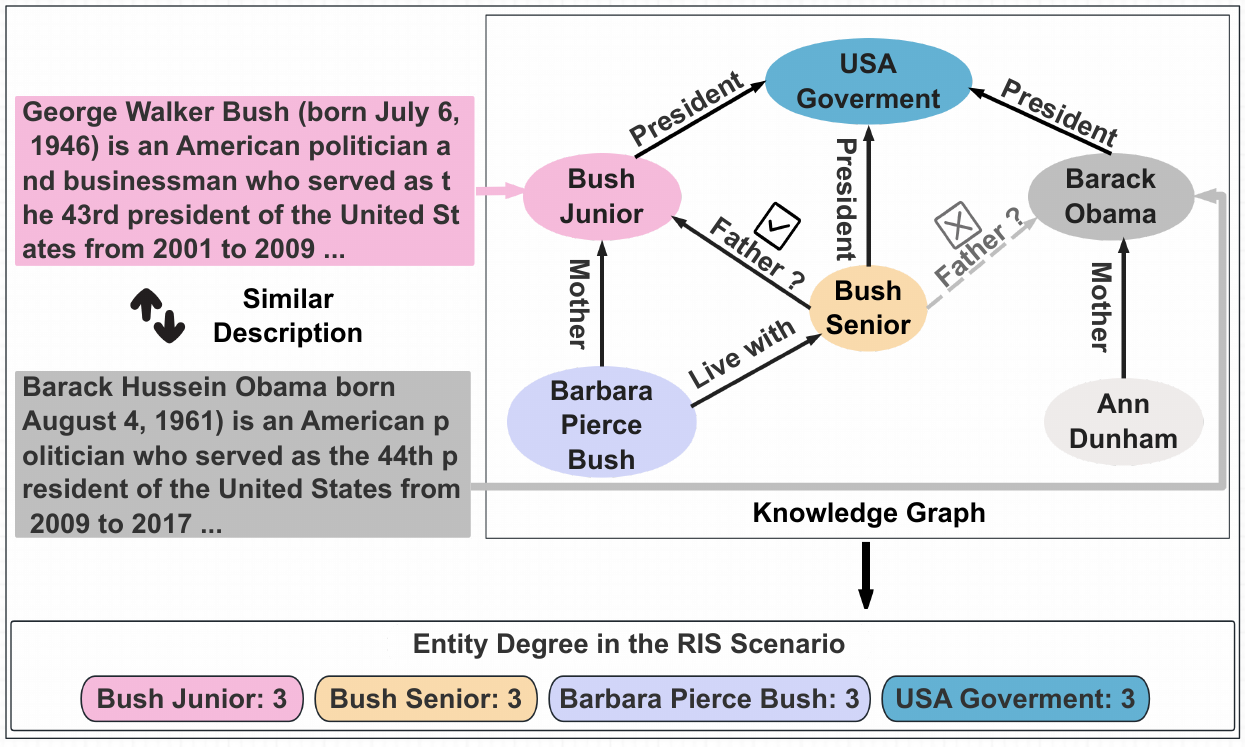}
            }
    \caption{\label{fig:1}
    Two Example Cases of Relation Prediction. Entity degree is the max of the sum of the out-degree and in-degree, or the number of paths from this entity to the connected entities. $\text{Entity Degree} = {max\{\text {(in-degree + out-degree), paths\}}}$.
    }
    }
    \vspace{-8mm}
\end{figure}
Knowledge Graph (KG) is a structured representation of the triplet~\cite{bordes2011learning,fusion2,ali}. 
However, in real-world scenarios, the issue of incomplete triples exists in most KGs~\cite{kgc_review_1,kgc_review_2}.

Existing Knowledge Graph Completion (KGC) methods can be divided into two main classes: single-knowledge-based models~\cite{transe,neurip,Random}, and multi-knowledge-fusion-based models~\cite{zhang2020relational,kgbert,pathcon,doloers}.
For most single-knowledge-based KGC models, such as TransE~\cite{transe}, TransH~\cite{transh}, TransD~\cite{transd}, and TransR~\cite{transr}, always depend on specific features within the KG. They primarily utilize embeddings of head and tail entities and calculate scores for potential relation candidates using corresponding translation functions, selecting the highest-scoring candidate for relation prediction. 
Since many paths in the KG contain more than two entities from the head entity to the end entity, some studies have concentrated on the path ranking algorithms \cite{Random} and rule mining methods \cite{neurip,drum} to improve the search efficiency for these multi-entity paths.
Besides, inspired by the GNNs in sub-graph representation learning, some KGC methods adopt the node-based message passing mechanism to propagate and aggregate nodes' features~\cite{conv,inductive,semi}. 

Unlike traditional methods that focus on single feature learning, recent multi-knowledge-fusion-based KGC models explore the fusion of textual description and the graph structure~\cite{BERT-2019,fusion1,fusion2}, and the fusion of context messages and reasoning paths~\cite{pathcon}. 
Nevertheless, these two KGC models both suffer the long-tail problem in the entity and relation prediction task, especially when dealing with sparsely distributed graph nodes. This issue makes KGC tasks more challenging and leads to lower accuracy~\cite{longtail_v1}. 

In this paper, we introduce MUSE, a knowledge-aware reasoning model designed to predict missing relations by continuously training a specialized embedding space. MUSE employs a multi-knowledge reasoning mechanism encompassing \textit{Prior Knowledge Learning}, \textit{Context Message Passing}, and \textit{Relational Path Aggregation}. Specifically, during the prior knowledge learning, we apply BERT to encode the description of head/tail entities and fine-tune BERT through a relation classification task. 
Then we employ this fine-tuned BERT checkpoint to initial the graph and explore the sub-graph topology information for each given entity pair. Besides, MUSE aggregates the context messages through the relational edge passing.
Meanwhile, our model enhances the path representation by reasoning and concating the entities, and relations on each path.
As illustrated in Figure~\ref{fig:1a}, we inject the prior knowledge into the entity description when the context message is limited.
For the Rich Information Set (RIS) scenario in Figure~\ref{fig:1b}, the entity descriptions are highly similar and cannot predict the correct relationship. 
Therefore, we use the context messages and path knowledge to reason about missing relations.
The experimental results obtained on the NELL995 dataset demonstrate that MUSE outperforms the existing KGC models by more than \textbf{5.50\% H@1} and \textbf{4.20\% MRR} in the relation prediction task. 
Further analysis reveals that MUSE provides an effective multi-knowledge reasoning mechanism that can effectively and accurately enhance the representation of knowledge graphs.
\vspace{-2mm}
\section{Methodology}
\label{sec:Methodology}
\begin{figure}[!t]
\centerline{\includegraphics[width=0.45\textwidth]{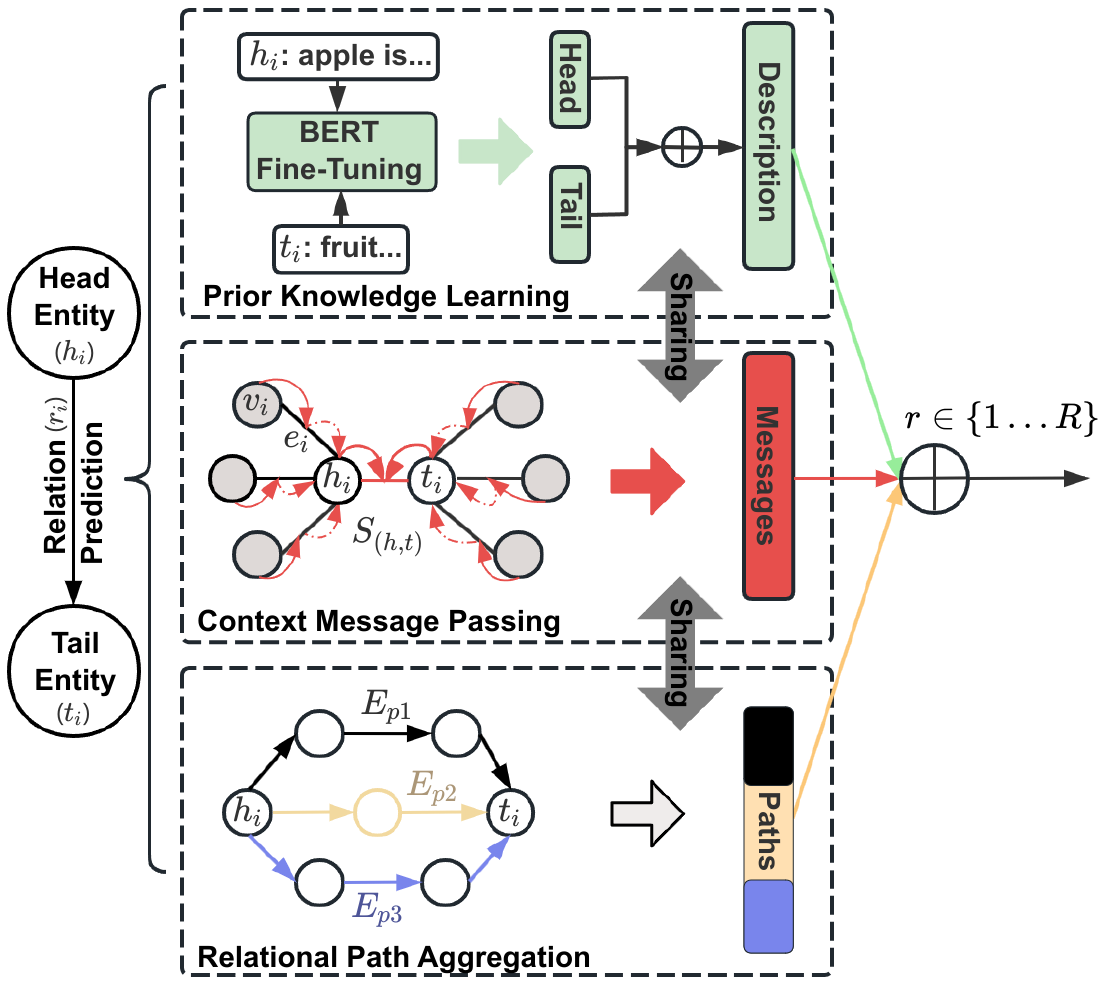}}
\caption{\label{1}The Architecture of MUSE Framework.}
\end{figure}
In this section, we present MUSE's framework. 
\subsection{Preliminary of MUSE}
As shown in Figure~\ref{1}, MUSE improves the prediction performance through: Prior Knowledge Learning; Context Message Passing; Relational Path Aggregation. Specifically, we need to predict the relation (${r}$) according to the head entity (\textbf{${h}$}), tail entity (\textbf{${t}$}). 
In the graph (\textbf{${G}$}), we also note the entity as node (${v}$) and the relation as edge (${e}$).
\subsubsection{Prior Knowledge Learning.\label{p}}
\begin{figure}[t]        
    \centering
\includegraphics[width=0.3\textwidth]{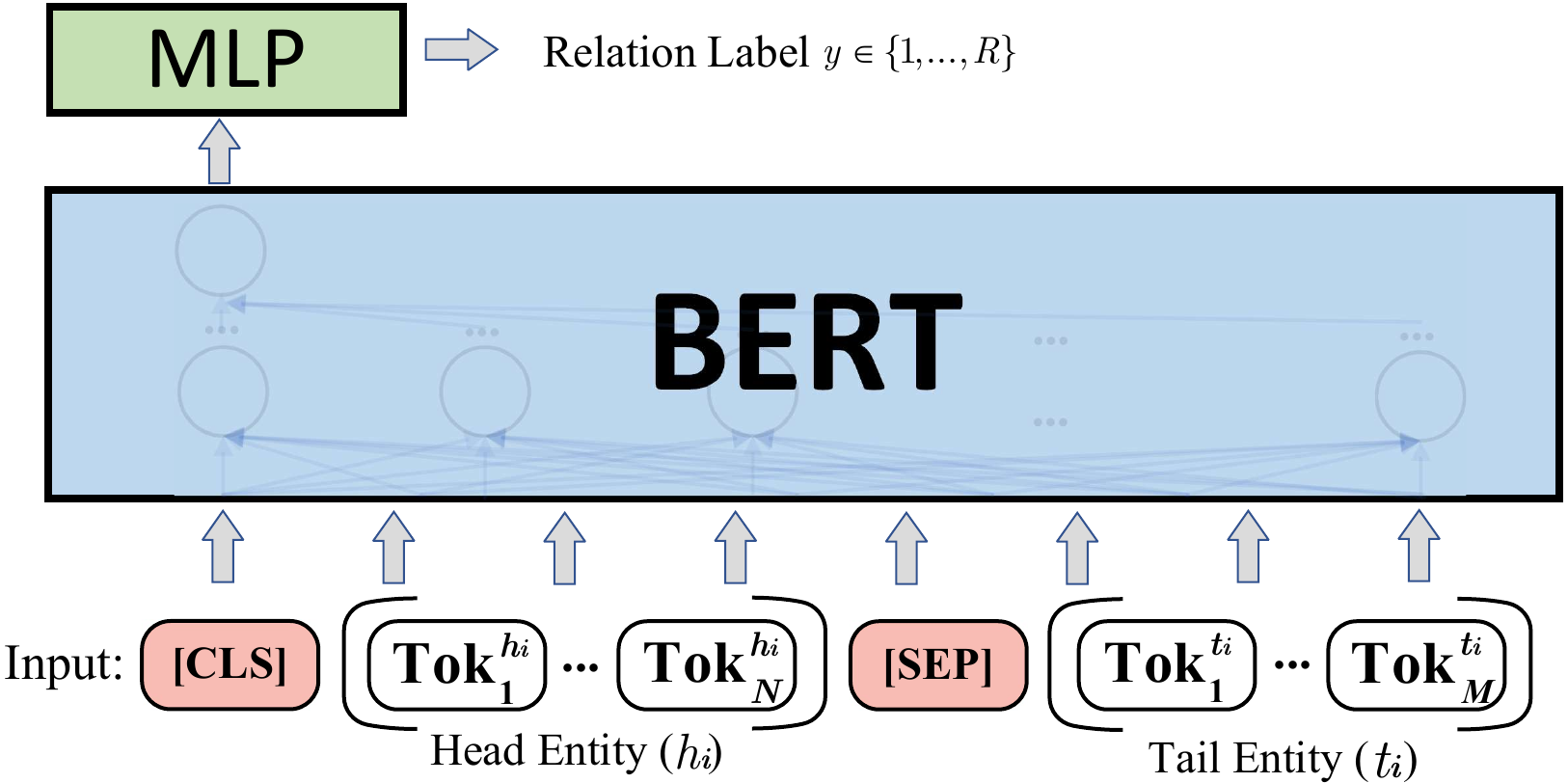} 
\caption{\label{fig:pkp}Illustration of the Prior Knowledge Learning. We fine-tune BERT on FB15k-237, WN18, WN18RR, and NELL995, respectively.}
\end{figure}

There exists some semantic knowledge hidden in the description of the entity. 
As shown in Figure~\ref{fig:pkp}, given the head entity (${h_{i}}$) and tail entity (${t_{i}}$), MUSE tokenises the words of entities as ${D_{hi}=\left\{\operatorname{Tok}_{1}^{h_{i}}, \ldots, \operatorname{Tok}_{N}^{h_{i}}\right\}}$ and ${D_{t i}=\left\{\operatorname{Tok}_{1}^{t_{i}}, \ldots, \operatorname{Tok}_{M}^{t_{i}}\right\}}$, respectively. 
Then our model combines the description set of these two entities (${E_{i}}$) as:
\begin{equation}
\small{{E}_{i}=\left\{[\text{CLS}], D_{hi}, [\text{SEP}], D_{ti}, [\text{SEP}]\right\}}
\end{equation}
MUSE employs BERT~\cite{BERT-2019} to encode this description set and takes the [CLS] token as the final hidden state (${{C_{i}}}$) to calculate the score (${S_{\tau i}}$)
\begin{equation}\label{eq:ft}
\small{{C}_{i}={BERT}({E}_{i})_{[CLS]},}
\end{equation}
\begin{equation}
\small{{S_{\tau i}= \mathrm{softmax} ({C_{i}}{W^{T}})},}
\end{equation}
where \( W \) represents the learnable weights in the classification layer. We then apply the triplet score and the true relation labels to calculate the relation classification task loss \( \mathcal{L}_{ft} \) as:
\begin{equation}\label{eq:input}
\small{{\mathcal{L}_{ft}=-\sum_{\tau \in \mathbb{G}} \sum_{j=1}^{R} y_{\tau i}^{j*} \log \left(S_{\tau i}^{j}\right)},}
\end{equation}
where ${y_{\tau i}^{j*}}$ denotes the relation indicator of triple, and ${j}$ represents any relation from $\{1, \ldots, R\}$. Specifically, for ${j = r}$, we have ${y_{\tau i}^{j*} = 1}$, otherwise if the ${j \neq r}$, we define ${y_{\tau i}^{j*} \neq 1}$.
\subsubsection{Context Message Passing.\label{c}}
MUSE follows the Pathcon~\cite{pathcon} model and designs an edge-based message passing mechanism to further enhance the representation of sub-graphs. Given the edge ${{e_{i}}}$ and node ${{v_{i}}}$, we can obtain the node's message representation ${{m_{v i}}}$ as: 
\begin{equation}\label{eq:attention}
{m_{vi}=\sum_{e \in \mathcal{N}(v)} \alpha_{ei} s_{ei}},
\end{equation}
\begin{equation}
\small{{\alpha_{e}=\frac{\exp \left(s_{e}^{T}\cdot  BERT{(v_{i})}\right)}{\sum_{e \in \mathcal{N}(v)} \exp \left(s_{e}^{T} \cdot  BERT{(v_{i})}\right)},}}
\end{equation}
where ${e \in \mathcal{N}(v)}$ represents the set of connected nodes. We employ the fine-tuned BERT checkpoint in Equation~\ref{eq:ft} to initialize their representation. Similarly, we can obtain the representation of the edge's message ${m_{ei}=\sum_{e \in \mathcal{N}(v)} \alpha_{vi} s_{vi}}$. The context representation ${s_{ei}}$ can be aggregated by the relation message passing:
\begin{equation}\label{eq:Sei}
\small{{S_{ei}=\sigma\left(\left[m_{ei} \| m_{vi} \| S_{e i-1}\right] \cdot W_{i}+b_{i}\right)},}
\end{equation}
where $\|$, ${W_{i}}$, $b_{i}$ and ${\sigma(\cdot) }$ denote the concatenation function, learnable transformation matrix, bias, and ${Relu}$ activation function. Then we can get the context message representation ${S_{(h, t)}}$ of entities after $K$ times message passing:
\begin{equation}\label{eq:input}
\small{{S_{(h, t)}=\sigma\left(\left[m_{hi}^{K} \| m_{ti}^{K}\right] \cdot W^{K}+b^{K}\right)}.}
\end{equation}
\begin{table}[t]
\centering
\resizebox{0.85\linewidth}{!}{
\small
\begin{tabular}{lrrrr} 
\hline
                                         & \textbf{FB15k-237}   & \textbf{WN18}        & \textbf{WN18RR}      & \textbf{NELL995}      \\ 
\hline
\multicolumn{1}{l}{\textbf{Raw Dataset}} & \multicolumn{1}{l}{} & \multicolumn{1}{l}{} & \multicolumn{1}{l}{} & \multicolumn{1}{l}{}  \\ 
\hline
Relation Type                          & 237                  & 18                   & 11                   & 198                   \\
 Entity Type                              & 14541               & 40943               & 40943               & 63917                \\
 Entity Degree Expectation                & 37.4                   & 6.9                    & 4.2                    & 4.3                     \\
 Entity Degree Variance                   & 12336.0               & 236.4                  & 64.3                   & 750.6                   \\ 
\hline
\multicolumn{1}{l}{\textbf{Data Splits}} & \multicolumn{1}{l}{} & \multicolumn{1}{l}{} & \multicolumn{1}{l}{} & \multicolumn{1}{l}{}  \\ 
\hline
Triplets in Train                        & 272115              & 141442              & 86835               & 137465               \\
Triplets in Development                 & 17535               & 5000                & 3034                & 5000                 \\
Triplets in Test                          & 20466               & 5000                & 3134                & 5000                 \\ 
\hline
\multicolumn{1}{l}{\textbf{Testing Scenario Percentages}}   & \multicolumn{1}{l}{} & \multicolumn{1}{l}{} & \multicolumn{1}{l}{} & \multicolumn{1}{l}{}  \\ 
\hline
Limited Information Set (\%)~            & 2                    & 7                    & 21                   & 31                    \\
Rich Information Set~(\%)~               & 98                   & 93                   & 79                   & 69                    \\
\hline
\end{tabular}
}
 \captionsetup{size=small,skip=5pt}
\caption{The Statistics Details in Our Experiments.}
\label{tab:statistics}
\end{table}
\subsubsection{Relational Path Aggregation.\label{r}}
For some similar entities in the KGC task, it is still difficult to predict the relation between the given head and tail entities relying on context knowledge \cite{pathcon}. Therefore, our model highlights the importance of identifying and capturing the relational paths between the given entity pairs.
Specifically, MUSE first uses the one-hot encoder to initial each path's representation ${E_{P}}$. Then we apply the context knowledge representation ${S_{ei}}$ in Equation~\ref{eq:Sei} to calculate the attention score ${\alpha_{P}}$ of this triple:
\begin{equation}
\small{{{\alpha}_{P}=\frac{\exp \left(E_{P}^{\top} \cdot S_{ei}\right)}{\sum_{P \in \mathcal{P}_{h \rightarrow t}} \exp \left(E_{P}^{\top} \cdot S_{ei}\right)}},}
\end{equation}
where the path set (${\mathcal{P}_{h \rightarrow t}}$) contains all the paths from the head entity to the tail entity. Then we can update the path knowledge representation ($S_{h \rightarrow t}$) as:
\begin{equation}
\small{{S_{h \rightarrow t}=\sum_{P \in \mathcal{P}_{h \rightarrow t}} \alpha_{P} E_{P}}.}
\end{equation}
\subsection{Multi-Knowledge Fusion}
After learning the semantic representation $S_{\tau_i}$ of triple \textbf{${\tau}$}, context message representation $S_{(h,t)}$ and the relational path knowledge $S_{h \rightarrow t}$. MUSE can achieve multi-knowledge fusion by aligning three knowledge representation learning:
\begin{equation}
\small{P(r \mid h, t) = \operatorname{softmax}\left(S_{\tau i} + S_{(h,t)} + S_{h \rightarrow t}\right),}
\end{equation}
where ${P(r\mid h, t)}$ is the probability of predicting the correct relation ${r}$ with given head and tail entities. MUSE then can be trained by minimizing the cross-entropy loss $\mathcal{L_{\tau}}$ as: 
{
\begin{equation}
\small{\mathcal{L_{\tau}}=\sum_{(h, r, t) \in \mathcal{T}} \operatorname{CrossEntropy}(r, P(r \mid h, t)).}
\end{equation}
}

\section{Experimental Methodology}
\label{sec:Experiment}
\begin{table*}[t]
\centering
\resizebox{0.9\linewidth}{!}{
\small
\begin{tabular}{l|ccc|ccc|ccc|ccc} 
\hline
\multicolumn{1}{l|}{\multirow{2}{*}{Models}} & \multicolumn{3}{c|}{FB15k-237}                                                                       & \multicolumn{3}{c|}{WN18}                                                                   & \multicolumn{3}{c|}{WN18RR}                                                                 & \multicolumn{3}{c}{NELL995}                                                                 \\ 
\cline{2-13}
\multicolumn{1}{c|}{}                         & MRR                                   & H@1                          & H@3                           & MRR                          & H@1                          & H@3                           & MRR                          & H@1                          & H@3                           & MRR                          & H@1                          & H@3                           \\ 
\hline
TransE                                        & 0.966                                 & 0.946                        & 0.984                         & 0.971                        & 0.955                        & 0.984                         & 0.784                        & 0.669                        & 0.870                         & 0.841                        & 0.781                        & 0.889                         \\
RotatE                                        & 0.970                                 & 0.951                        & 0.980                         & 0.984                        & 0.979\uline{}                & 0.986\uline{}                 & 0.799                        & 0.735                        & 0.823                         & 0.729                        & 0.691                        & 0.756                         \\
QuatE                                         & 0.974\uline{}                         & 0.958\uline{}                & 0.988\uline{}                 & 0.981                        & 0.975                        & 0.983                         & 0.823                        & 0.767                        & 0.852                         & 0.752\uline{}                & 0.706\uline{}                & 0.783\uline{}                 \\
DRUM                                          & 0.959                                 & 0.905                        & 0.958                         & 0.969                        & 0.956                        & 0.980                         & 0.854\uline{}                & 0.778\uline{}                & 0.912\uline{}                 & 0.715                        & 0.640                        & 0.740                         \\
\text{PathCon}$^{\spadesuit}$                                       & \uline{0.979}                         & \uline{0.964}                & \uline{0.994}                 & \uline{0.993}                & \uline{0.988}                & \uline{0.998}                 & \uline{0.974}                & \uline{0.954}                & 0.994                         & 0.896                        & \uline{0.844}                & 0.941                         \\
KG-BERT                                       & 0.973                                 & 0.953                        & 0.993                         & 0.992                        & 0.987                        & 0.997                         & \textbf{0.991}               & \textbf{0.983}               & \uline{0.999}                 & \uline{0.897}                & 0.821                        & \uline{0.970}                 \\ 
\hline
\multirow{2}{*}{\textbf{MUSE}}                & \textbf{0.985}                        & \textbf{0.974}               & \textbf{0.997}                & \textbf{0.995}               & \textbf{0.992}               & \textbf{1.000}                & 0.986                        & 0.975                        & \textbf{1.000}                & \textbf{0.939}               & \textbf{0.899}               & \textbf{0.981}                \\
                                              & \multicolumn{1}{l}{\textbf{± }0.000} & \multicolumn{1}{l}{± 0.001} & \multicolumn{1}{l|}{± 0.000} & \multicolumn{1}{l}{± 0.001} & \multicolumn{1}{l}{± 0.001} & \multicolumn{1}{l|}{± 0.000} & \multicolumn{1}{l}{± 0.001} & \multicolumn{1}{l}{± 0.002} & \multicolumn{1}{l|}{± 0.000} & \multicolumn{1}{l}{± 0.002} & \multicolumn{1}{l}{± 0.003} & \multicolumn{1}{l}{± 0.002}  \\
\hline
\end{tabular}
}
\captionsetup{size=small,skip=5pt}
\caption{Relation Prediction in the General Scenario. The best results are highlighted in \textbf{bold}, and the best results of the baseline are \uline{underlined}. Besides, the output results of \text{PathCon}$^{\spadesuit}$ are from the paper of Wang et al~\cite{pathcon}. Our experiment is repeated three times and we report the average results with the corresponding standard deviation.}
\label{tab:overall}
\end{table*}
In this section, the experimental settings of our model, MUSE, and other baseline models are described.
\subsection{Datasets}
We conduct evaluations of MUSE on four public datasets widely used in Knowledge Graph Completion (KGC) task: FB15k-237~\cite{observed}, WN18~\cite{bordes2011learning}, WN18RR~\cite{Dettmers_Minervini_Stenetorp_Riedel_2018}, and NELL995~\cite{nell}. More details of statistics are list
in Table~\ref{tab:statistics}. 
We observe the expectation and variance of entity degree are quite different across our four datasets, for example, the expectation is 4.2 and the variance is 750.6 on the NELL995, while these two statistics are 37.4 and 12336 on the FB15k-237.

\textit{Testing Scenarios}.
We have established the \textbf{Limited Information Set (LIS)} scenario and \textbf{Rich Information Set (RIS)} scenario according to the entity degree in the Knowledge Graph (KG).
Specifically, LIS consists of entities with degrees lower than three, and RIS includes entities with degrees equal/higher than three.
Besides, the degree of an entity is the max of the sum of the out-degree and in-degree, or the number of paths.
\subsection{Baselines}
We apply several typical KGC models as baselines to compare with our model in the relation prediction task. 

\textit{Single-Knowledge-Based models}: TransE~\cite{transe}, RotatE~\cite{rotate}, and QuatE~\cite{quate} are embedding-based methods. The main difference among them is the type of continuous space for entities. DRUM~\cite{drum} is a path representation learning method capturing path features of entities using probabilistic logical rules.
  
\textit{Multi-Knowledge-Fusion-Based models}: PathCon~\cite{pathcon} is one of the latest SOTA KGC methods, which can learn both the context information and relational path features from the head entity to the target tail entity. KG-BERT~\cite{kgbert} is a method that enhances textual features of entities by fine-tuning the BERT.

\textit{Evaluation Metrics}.
The official KGC evaluation metrics include Mean Reciprocal Rank (MRR), HIT@1 (H@1), and HIT@3 (H@3). The H@1 is our main evaluation.
\subsection{Implementation Details}
\vspace{-2mm}
For TransE, RotatE, QuatE, and DRUM, we set the embedding dimensions at 400 and training epoch is 1000. And we follow the hyper-parameter setting of KG-BERT\footnote{\url{ https://github.com/yao8839836/kg-bert}} and PathCon\footnote{\url{https://github.com/hwwang55/PathCon}}.

For the MUSE implementation, we start from the Bert-base-uncased and fine-tunes this language model using prior knowledge learning. 
Specifically, during BERT model embedding, we set the max length of each entity description to 512 tokens and fine-tuned it for 10 training epochs.
MUSE then follows the optimal parameter setting used by PathCon~\cite{pathcon} to extract contextual and path features.
In our experiments, the specific configuration
of the [FB15k-237, WN18, WN18RR, NELL995] dataset is: the number of context layers is set to [2, 3, 3, 2], the max path length is [3, 3, 4, 5], the learning rate is set to [1e-4, 1e-4, 5e-4, 1e-4]. 
Besides, we use the Adam optimizer and set the batch size to 128, the training epoch to 60. 
All experiments are conducted on 2 NVIDIA NTX 3090ti GPUs.
\section{Evaluation Results}
In this section, the experimental settings of our model, MUSE and other baseline models are described.
\begin{table}[t]
\small
\centering
\resizebox{\linewidth}{!}{
\begin{tabular}{l|c|c|c|c} 
\hline
\multirow{3}{*}{Methods} & \multicolumn{4}{c}{Datasets}                                                                              \\ 
\cline{2-5}
                         & FB15k-237               & WN18                       & WN18RR                  & NELL995                  \\ 
\cline{2-5}
                         & H@1                     & H@1                        & H@1                     & H@1                      \\ 
\hline
MUSE (Full Model)        & \textbf{\textbf{0.974}} & \textbf{\textbf{0.992}}    & \textbf{\textbf{0.975}} & \textbf{\textbf{0.899}}  \\ 
\hline
w/o~ Prior Knowledge     & 0.964                   & 0.988                      & 0.954                   & 0.844                    \\
w/o~ Context Message     & 0.973                   & 0.988                      & 0.965                   & 0.892                    \\
w/o~ Relational Path     & 0.965                   & 0.954                      & 0.903                   & 0.874                    \\
\hline
w/o~ All (Backbone Model)      & 0.943                   & 0.951   & 0.661                   & 0.779                    \\
\hline
\end{tabular}
 }
\captionsetup{size=small,skip=5pt}
\caption{Ablation Study in the Relation Prediction Task.}
\label{tab:as}
\end{table}
\vspace{-2mm}
\subsection{Overall Performance}
This subsection evaluates the results of MUSE and baselines in the relation prediction task. We find that our model has almost achieved the best performance across all datasets, including the FB15K-237, WN18, WN18RR, and NELL995.

As shown in Table \ref{tab:overall}, MUSE improves H@1 by 1\%, 0.4\%, and 5.5\% over baseline models on the FB15K-237, WN18, and NELL995 datasets, respectively.
Besides, our model has already achieved 1.0 accuracy at H@3 on both WN18 and WN18RR datasets, and H@3 on NELL995 steadily improves by 4.0\% and 1.1\% compared with PathCon and KG-BERT.
Notably, MUSE performs better in the Knowledge Graph (KG) with more sparsely distributed nodes. Specifically, our model achieves the most significant increase compared to PathCon on the WN18RR and NELL995 datasets, which have the highest Limited Information Set (LIS) proportion over 21\% and 31\%. 
\subsection{Ablation Study}
MUSE has three parallel components: \textit{Prior Knowledge Learning}, \textit{Context Message Passing}, and \textit{Relational Path Aggregation}. 
In Table~\ref{tab:as}, we conduct the ablation study to investigate the role played by each representation learning module. The backbone model randomly initializes the knowledge graph.

Generally, we observe that MUSE (full model) exceeds all ablation models when adopting the multi-knowledge representation learning mechanism. For higher LIS proportion datasets, including WN18RR and NELL995, MUSE grains more improvement (0.314 and 0.12) compared to the backbone model.

In addition, the semantic features in prior knowledge effectively direct MUSE toward the correct relation. Specifically, the prior knowledge module can cause significant (0.055) decrease on the NELL995 dataset.
Then MUSE mainly acquires topological features in the knowledge graph with more RIS scenarios.
Specifically, the prediction performance of the w/o relational path model is 0.954 H@1 on the FB15k-237 dataset, while the backbone model can reach 0.951. This supports our research that injecting the semantic knowledge to the graph can effectively enrich the representation of entities, and significantly improve the relation prediction accuracy. 
\begin{figure}[t!]
    \centering 
    \subfigure[The Effectiveness of Fine-Tuning the BERT Model.] { 
    \label{fig:pkla} 
     \includegraphics[width=0.4\linewidth]{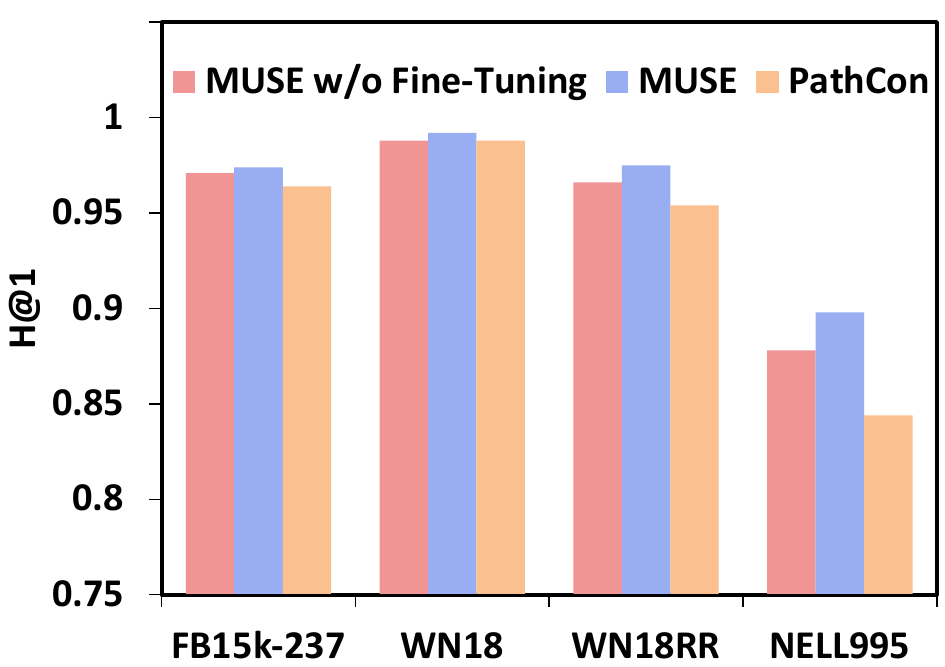}}
     \subfigure[The Effectiveness of Edge-Based Attention Mechanism.] { 
     \label{fig:pklb}
    \includegraphics[width=0.4\linewidth]{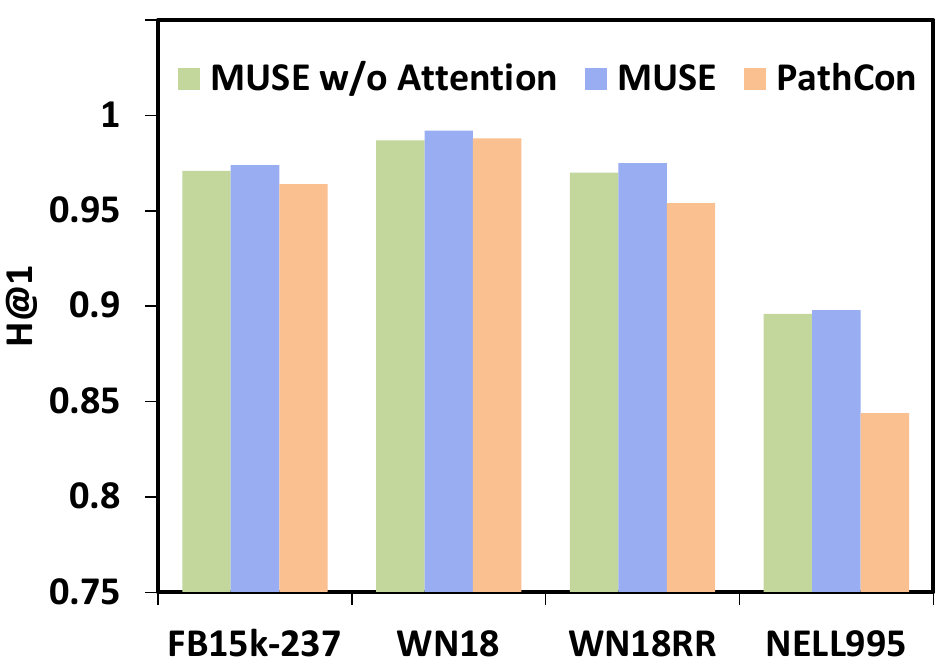}}
    \caption{Analysis of the External Semantic Knowledge in the Relation Prediction Task. 
    We define the $\text{\textbf{{MUSE~w/o~Fine-Tuning}}}$ model as applying the BERT model directly in Figure \ref{fig:pkla}. 
   As shown in Figure \ref{fig:pklb}, the $\text{\textbf{{MUSE w/o Attention}}}$ model aggregates entities without using attention mechanism in Equation \ref{eq:attention}.}
    \label{fig:pkl}
    \vspace{-2mm}
\end{figure}

\begin{figure}[t]
    \centering 
    \subfigure[Relation Prediction under Different Number of Paths.] { 
    \label{fig:4a} 
     \includegraphics[width=0.4\linewidth]{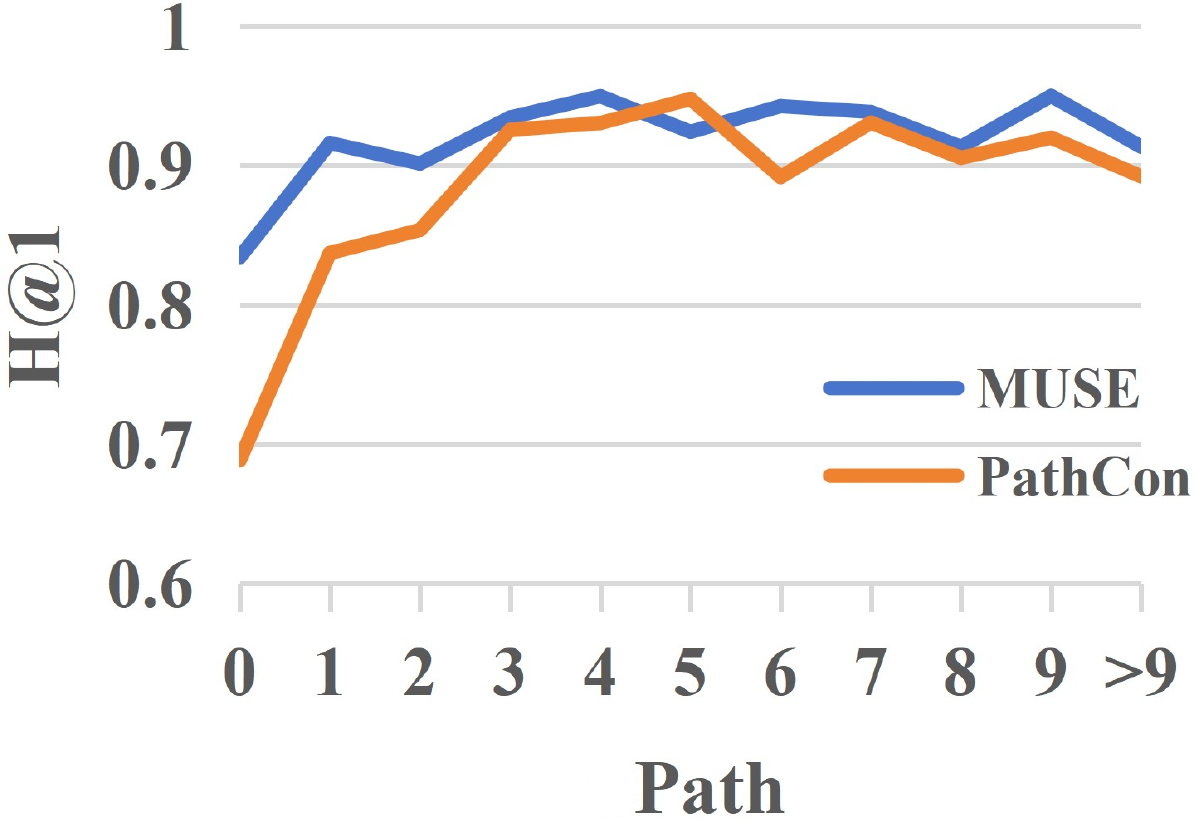}}
     \hspace{-1mm}
     \subfigure[Relation Prediction under Different Number of Degrees.] { 
     \label{fig:4b}
    \includegraphics[width=0.4\linewidth]{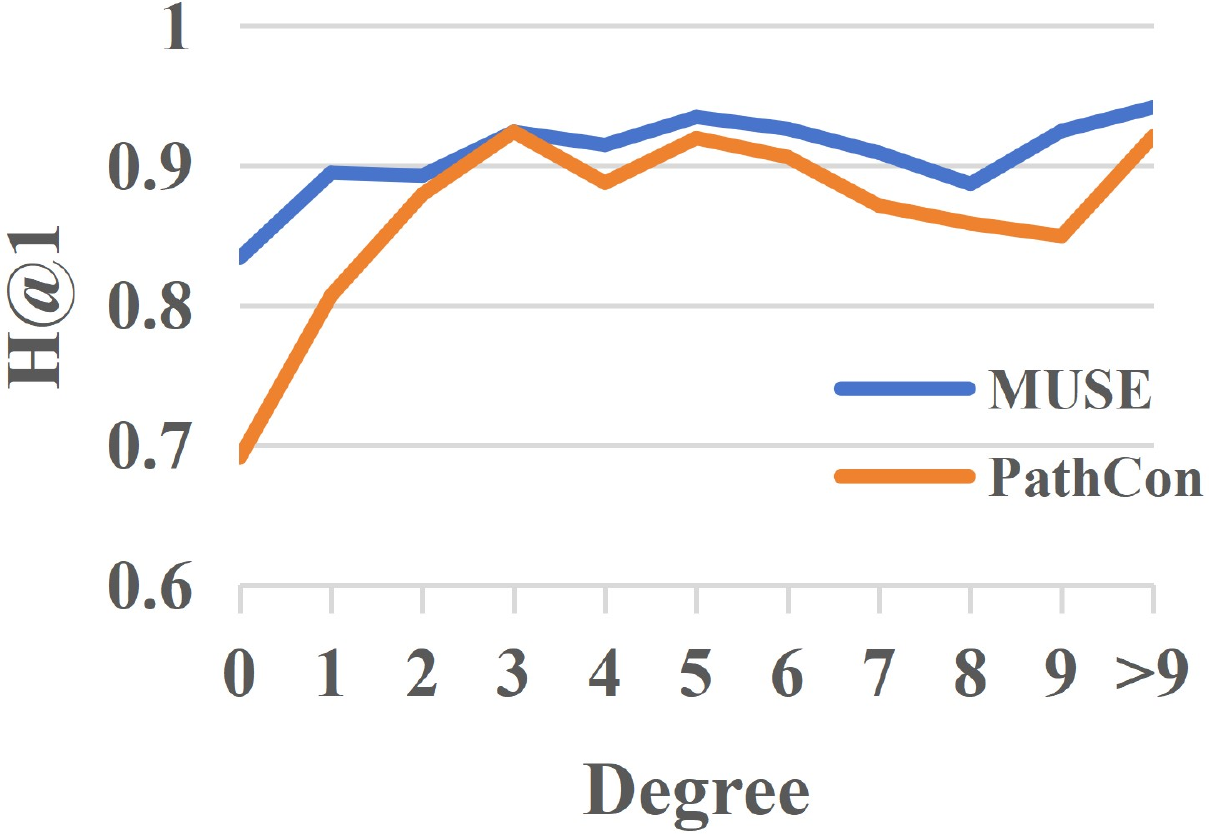}}
    \caption{Performance of MUSE and PathCon on the NELL995.}
    \label{fig:line}
\end{figure}

\subsection{Contributions of Multi-Knowledge in MUSE}
In this subsection, we investigate the effectiveness of MUSE (our model) and PathCon (best baseline model) on all datasets.

\textit{Guidance of Semantic Knowledge Injection} 

Figure \ref{fig:pkla} shows that MUSE consistently exceeds the performance of PathCon on all datasets after BERT initializes the graph.
Additionally, using the fine-tuned BERT can further enhance the representations of entities effectively and accurately, achieving a 0.034 H@1 improvement on the NELL995 dataset.

Another analysis presented in Figure~\ref{fig:pklb} concentrates on the edge-based message passing mechanism.
Similarly, the experiments show that MUSE on each dataset is higher than the model without the attention mechanism. It proves the effectiveness of semantics interaction in relation prediction. 
Besides, compared to these two semantic knowledge injection strategies, fine-tuning the language model can guide MUSE to learn sufficient entity representation and improve performance.

\textit{Effectiveness of Entity Semantic Representation.} 

External semantic knowledge plays a crucial role in entity representation and improves prediction accuracy even in graph-structured data-rich environments. Our experiments further analysis the impact of the prior knowledge learning on the NELL995 dataset. We compare the performance of MUSE and PathCon across various triplet paths and degrees in Figure~\ref{fig:4a} and Figure~\ref{fig:4b}, respectively. 
The results show that our model consistently outperforms PathCon in all experiments, especially when both paths and degrees are less than three.

\section{Conclusion}
\vspace{-2mm}
\label{sec:Conclusion}
We present MUSE, a sophisticated model combining \textit{Prior Knowledge Learning}, \textit{Context Message Passing}, and \textit{Relational Path Aggregation} to advance entity representation in relation prediction.
Ablation studies and semantic evaluations validate the significance in various knowledge graph applications.
\bibliographystyle{splncs04}
\bibliography{refs}

\begin{thebibliography}{10}
\providecommand{\url}[1]{\texttt{#1}}
\providecommand{\urlprefix}{URL }
\providecommand{\doi}[1]{https://doi.org/#1}

\bibitem{transe}
Bordes, A., Usunier, N., Garcia-Duran, A., Weston, J., Yakhnenko, O.: Translating embeddings for modeling multi-relational data  (2013)

\bibitem{bordes2011learning}
Bordes, A., Weston, J., Collobert, R., Bengio, Y.: Learning structured embeddings of knowledge bases. In: Proceedings of AAAI. pp. 301--306 (2011)

\bibitem{Dettmers_Minervini_Stenetorp_Riedel_2018}
Dettmers, T., Minervini, P., Stenetorp, P., Riedel, S.: Convolutional 2d knowledge graph embeddings  (2018)

\bibitem{BERT-2019}
Devlin, J., Toutanova, K.: {BERT}: Pre-training of deep bidirectional transformers for language understanding. In: Proceedings of NAACL-HLT (2019)

\bibitem{conv}
Duvenaud, D., Adams, R.P.: Convolutional networks on graphs for learning molecular fingerprints (2015)

\bibitem{inductive}
Hamilton, W.L., Ying, R., Leskovec, J.: Inductive representation learning on large graphs (2018)

\bibitem{transd}
Ji, G., He, S., Xu, L., Liu, K., Zhao, J.: Knowledge graph embedding via dynamic mapping matrix. In: Proceedings of ACL-IJCNLP. pp. 687--696 (2015)

\bibitem{semi}
Kipf, T.N., Welling, M.: Semi-supervised classification with graph convolutional networks. arXiv  (2016)

\bibitem{Random}
Lao, N., Mitchell, T., Cohen, W.: Random walk inference and learning in a large scale knowledge base. In: Proceedings of EMNLP. pp. 529--539 (2011)

\bibitem{transr}
Lin, Y., Liu, Z., Sun, M.: Learning entity and relation embeddings for knowledge graph completion. In: Proceedings of AAAI (2015)

\bibitem{drum}
Sadeghian, A., Armandpour, M., Ding, P., Wang, D.Z.: Drum: End-to-end differentiable rule mining on knowledge graphs  (2019)

\bibitem{kgc_review_1}
Shen, T., Zhang, F., Cheng, J.: A comprehensive overview of knowledge graph completion. Knowledge-Based Systems p. 109597 (2022)

\bibitem{rotate}
Sun, Z., Deng, Z.H., Nie, J.Y., Tang, J.: Rotate: Knowledge graph embedding by relational rotation in complex space. arXiv preprint  (2019)

\bibitem{observed}
Toutanova, K., Chen, D.: Observed versus latent features for knowledge base and text inference. In: Proceedings of the Workshop on Continuous Vector Space Models and their Compositionality (2015)

\bibitem{fusion1}
Toutanova, K., Gamon, M.: Representing text for joint embedding of text and knowledge bases. In: Proceedings of EMNLP. pp. 1499--1509 (2015)

\bibitem{doloers}
Wang, H., Kulkarni, V., Wang, W.Y.: Dolores: Deep contextualized knowledge graph embeddings  (2018)

\bibitem{pathcon}
Wang, H., Ren, H., Leskovec, J.: Relational message passing for knowledge graph completion. In: Proceedings of KDD. pp. 1697--1707 (2021)

\bibitem{kgc_review_2}
Wang, M., Wang, X.: A survey on knowledge graph embeddings for link prediction. Symmetry p.~485 (2021)

\bibitem{transh}
Wang, Z., Zhang, J., Feng, J., Chen, Z.: Knowledge graph embedding by translating on hyperplanes. In: Proceedings of AAAI (2014)

\bibitem{fusion2}
Xie, R., Liu, Z., Jia, J., Luan, H., Sun, M.: Representation learning of knowledge graphs with entity descriptions. In: Proceedings of AAAI (2016)

\bibitem{nell}
Xiong, W., Hoang, T., Wang, W.Y.: Deeppath: A reinforcement learning method for knowledge graph reasoning. arXiv preprint {\color{blue}}  (2017)

\bibitem{neurip}
Yang, F., Yang, Z., Cohen, W.W.: Differentiable learning of logical rules for knowledge base reasoning

\bibitem{kgbert}
Yao, L., Mao, C., Luo, Y.: Kg-bert: Bert for knowledge graph completion. arXiv preprint  (2019)

\bibitem{ali}
Zeng, K., Li, J., Feng, L.: A comprehensive survey of entity alignment for knowledge graphs. AI Open  (2021)

\bibitem{longtail_v1}
Zhang, N., Chen, X., Zhang, W., Chen, H.: Long-tail relation extraction via knowledge graph embeddings and graph convolution networks. arXiv preprint  (2019)

\bibitem{quate}
Zhang, S., Tay, Y., Yao, L., Liu, Q.: Quaternion knowledge graph embeddings  \textbf{32} (2019)

\bibitem{zhang2020relational}
Zhang, Z., Xiong, H., He, Q.: Relational graph neural network with hierarchical attention for knowledge graph completion. In: Proceedings of AAAI (2020)

\end{thebibliography}



\end{document}